\documentclass[10pt,twocolumn,letterpaper]{article}

\usepackage{cvpr}
\usepackage{times}
\usepackage{epsfig}
\usepackage{graphicx}
\usepackage{amsmath}
\usepackage{amssymb}
\usepackage{algorithm}
\usepackage[noend]{algpseudocode}
\usepackage{graphicx}
\usepackage{color}
\usepackage{indentfirst}

\usepackage[pagebackref=true,breaklinks=true,letterpaper=true,colorlinks,bookmarks=false]{hyperref}

\def\cvprPaperID{2} 

\ifcvprfinal\fi
\begin{document}


\title{A method for detecting text of arbitrary shapes in natural scenes that improves text spotting}

\author{Qitong Wang, Yi Zheng, and Margrit Betke \\
Boston University \\
Boston, MA 02215 \\
{\tt\small \{wqt1996, yizheng, betke\}@bu.edu}}


\maketitle

\begin{abstract}
   
 Understanding the meaning of text in images of natural scenes like highway signs or store front emblems is particularly challenging if the text is foreshortened in the image or the letters are artistically distorted.  We introduce a  pipeline-based text spotting framework that can both detect and recognize text in various fonts, shapes, and orientations in natural scene images with complicated backgrounds. The main contribution of our work is the text detection component, which we call {\em UHT,} short for UNet, Heatmap, and Textfill.  UHT uses a
UNet to compute heatmaps for candidate text regions and a textfill algorithm to produce tight polygonal boundaries around each word in the candidate text. Our method trains the UNet with groundtruth heatmaps that we obtain from text bounding polygons provided by groundtruth annotations. 
Our text spotting framework, called UHTA, combines UHT with the state-of-the-art text recognition system ASTER.  
  Experiments on four challenging and public scene-text-detection datasets (Total-Text, SCUT-CTW1500, MSRA-TD500, and COCO-Text)
   show the effectiveness and generalization ability of UHT in detecting not only multilingual (potentially rotated) straight but also curved text in scripts of multiple languages. Our experimental results of UHTA on the Total-Text dataset show that UHTA outperforms four state-of-the-art text spotting frameworks 
   by at least 9.1 percent points in the F-measure, which suggests that UHTA may be used as a complete text detection and recognition system in real applications.
\end{abstract}



\section{Introduction}

\begin{figure}[t]
\centering
\centering
\includegraphics[width=1\columnwidth]{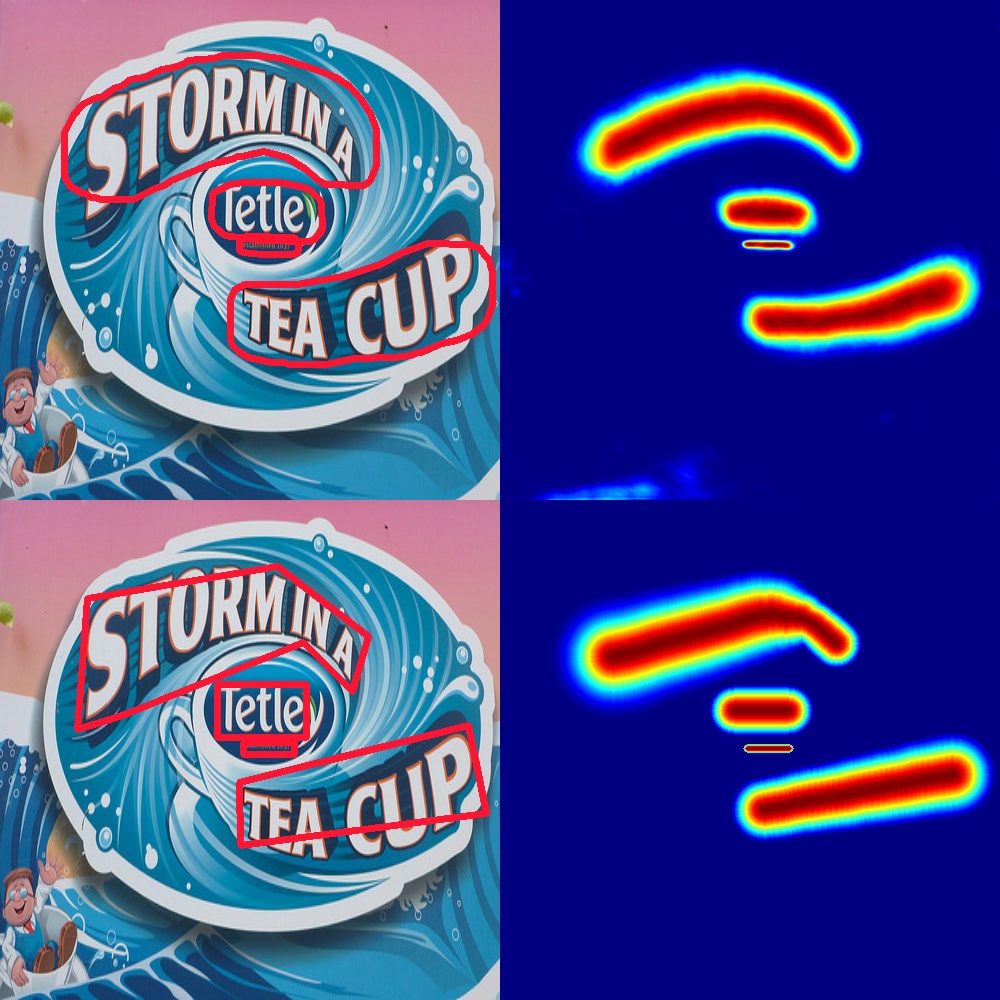}
\caption{
Polygonal text annotations of curved text in images are often so imprecise (bottom left) that heat maps (bottom right), computed as an intermediate step to interpret the text, yield inaccurate results.  The proposed UHT method computes and interprets deep learned heat maps (top right) that result in much more accurate polygonal text outlines (top left), which in turn yield better text recognition results.
}
\label{vis}
\end{figure}

Scene text detection is an important task in computer vision with application significance such as helping people with visual impairments to understand text in images (e.g., of medicine bottles or supermarket shelves) or helping self-driving cars understand the meaning of traffic and street signs.
Building computer vision systems that can detect text is not easy due to the variety of sizes, fonts, styles, sizes, and orientations in which text can occur in natural scene images and their often complex backgrounds (e.g., Fig.~\ref{vis}). 

In the past few years, computer vision researchers have developed methods that  identify oriented straight text in natural scene images accurately \cite{DBLP:journals/corr/HeZYL17, DBLP:journals/corr/abs-1802-08948, DBLP:journals/corr/YaoBSZZC16, 
DBLP:journals/corr/ZhangZSYLB16, east} (``oriented'' means not necessarily aligned with the image rows).  More recently, detection of arbitrarily-shaped text, such as curved or deformed text, has received attention from computer vision researchers, not only because detecting such text is more challenging than oriented straight text, but also because it commonly appears in daily life. 
For arbitrarily-shaped text detection, for example, a weakly supervised learning algorithm~\cite{craft} was recently proposed to extract character-based pseudo ground truth to help deep learning models effectively extract each character of such a text in a natural scene image.  Whether it is detection of oriented straight or curved text, we found that most state-of-the-art methods rely on multiple deep-learned geometric properties of the text,  
such as angle attributes \cite{textsnake, east} or text center line regions \cite{lomo-ms}. Others use multiple output models \cite{lomo-ms} to produce high evaluation scores on widely-used benchmarks.  
While state-of-the-art text detection methods can solve many challenging problems with these techniques, as far as we know, there is no method that simply and effectively uses a ``text region feature map,'' even when given a variety of text shapes, sizes, and lengths.  Moreover, many words are located so close to each other in the images that detection methods do not separate them correctly but grouped into one consecutive-word text region.  These challenges make relying on only the text region feature map to effectively detect text in natural scene images seemingly impossible. But is it really impossible to accurately detect text using only one text region feature map in the scene text detection field?  Our work shows that the answer is no.
Using only one channel, the text region feature map, 
our method effectively detects text in images of natural scenes. With the help of new pre-processing and post-processing algorithms, we make accurately detecting text in images possible, relying on a relatively small amount of geometric information. 

The contributions of our research work are five-fold:

$\bullet$ We propose a new text detection framework, called UHT, that outputs only one text region heatmap channel. 
UHT can solve challenging problems in the field of scene text detection, such as accurately detecting and separating multiple text regions that ``stick'' together.

$\bullet$ We propose a new text region feature map representation, which here is a special kind of heat map (Fig.~\ref{vis}), that enables UHT to detect text in natural scene image.

$\bullet$ We propose a new algorithm called the {\em Textfill Algorithm} that can accurately extract multi-vertex bounding polygons that tightly define the outline of each word in the scene text region. 


$\bullet$ UHT obtained evaluation scores that are higher than most of state-of-the-art scene text detection methods when fine-tuned on specific benchmark datasets. UHT outperforms all state-of-the-art methods in its generalization ability, as shown in one of the experiments.

$\bullet$ ``Spotting'' text in images means detecting and recognizing it.  We introduce a complete pipeline-based text spotting system, called UHTA, showing that our UHT can be used for text spotting as long as an effective text recognition model is given.

Our code is available at http://www.cs.bu.edu/faculty/ betke/UHT.

\begin{figure}[t]
\centering
\centering
\includegraphics[width=1\columnwidth]{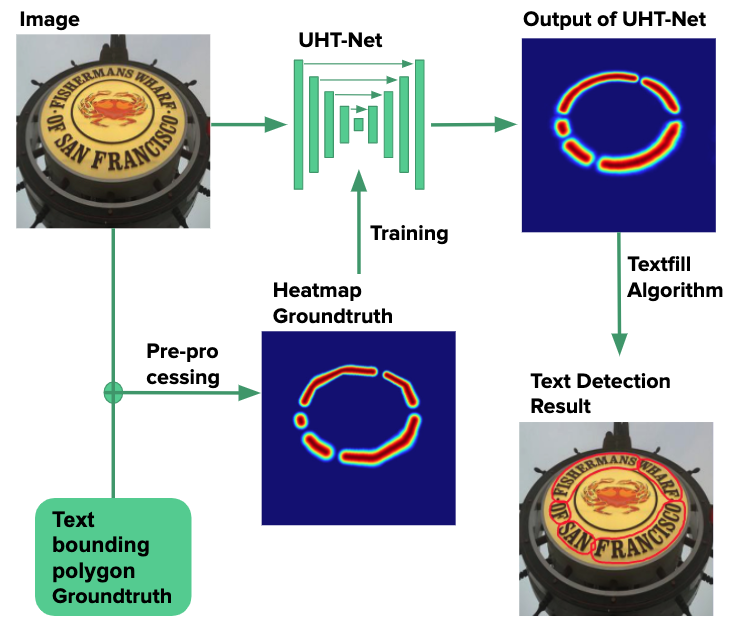}
\caption{Pipeline of UHT. The process of text detection of UHT can be divided into three steps: 1) Pre-processing is used to generate a heatmap text region ground truth, which is used as a training label of UHT-Net. 2) A trained UHT-Net can output predicted text region heatmaps. 3) In the post-processing step, the Textfill algorithm outputs the final predicted text bounding polygons interpreting the outputs of the UHT-Net.}
\label{pipeline}
\end{figure}

\section{Related Works}

The task of detecting text in images of everyday scenes, also known as "Scene Text Detection" is attracting more and more attention from researchers in the computer vision field. Initially, researchers focused on detecting oriented straight text in scene images \cite{east, pixellink, DBLP:journals/corr/ZhangZSYLB16, DBLP:journals/corr/YaoBSZZC16, DBLP:journals/corr/HeZYL17}. However, detecting text with arbitrary shapes is more and more popular recently \cite{textsnake, psenet-1s, psenet_v2, lomo-ms, craft, textmountaion, charnet}.

Before methodologies in deep learning field are widely used in text detection field, SWT \cite{swt} and MSER \cite{mser} were two eye-catching algorithms which had influenced many text detection methodologies. In recent years, modern methodologies, which make use of deep learning backbones, can be coarsely classified into two categories: regression-based methodologies and segmentation-based methodologies.

{\bf Regression-based methodologies} are largely influenced by some popular general object detection frameworks such as Faster-RCNN \cite{faster-rcnn}. TextBoxes \cite{textboxes} was inspired by SSD \cite{ssd} and included “long” default boxes that had large aspect ratios to better detect text with different variation in natural scene images. In the text detection branch of Mask-TextSpotter \cite{mask_textspotter}, many text proposals were firstly generated by region proposal network to get text candidate boxes, then the RoI features of the text proposals were sent into the Fast R-CNN module.

{\bf Segmentation-based methodologies} are mainly inspired by FCN \cite{fcn}, The FCN classifies the image at the pixel level, thus solving the problem of image segmentation at the semantic level. In the text detection field, people see text regions in natural scene images as positive samples and background as negative samples. TextSnake \cite{textsnake} was proposed to detect text in the natural scene by predicting the text region and various geometry attributes of text to detect oriented straight and curve text effectively. Recently, instead of detecting whole text in images, CRAFT \cite{craft} was proposed to detect individual characters, connecting them to get each text bounding polygon. The proposed method provides the character region score and the character affinity score that, together, effectively cover various kinds of text shapes. In this method, a weakly-supervised framework was implemented to generate character-level pseudo annotations.

As we can see, state-of-the-art frameworks make full use of a large volume of geometric information to effectively detect text in natural scene images. Our methodology, however, is based on only using text region information to effectively extract text bounding polygons from images.

\section{Methodology}

The pipeline of our model is shown in Figure~\ref{pipeline}.  We now introduce our methodology in detail.

\subsection{Pre-processing: Heatmap Text Region Groundtruth Generation}

\begin{figure}[tb]
\centering
\centering
\includegraphics[width=1.00\columnwidth]{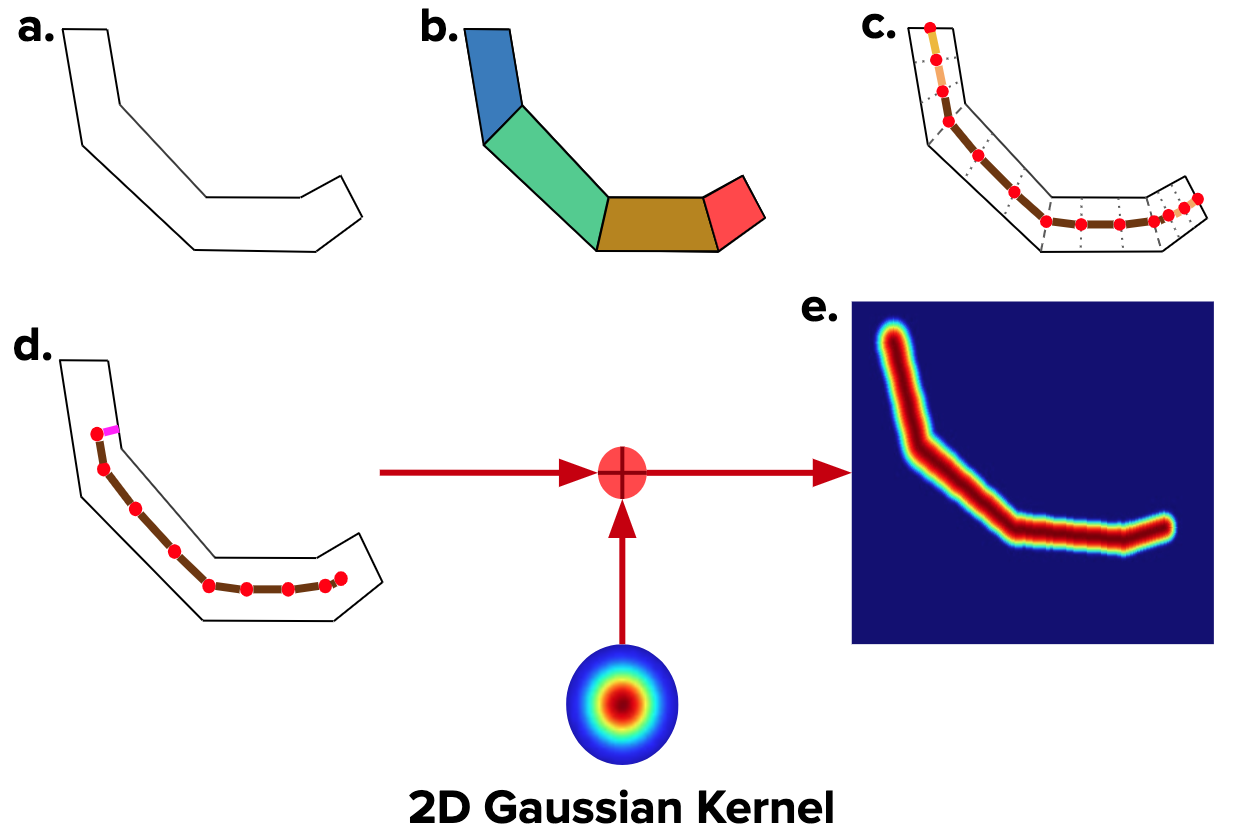}
\caption{Process of creating the heatmap groundtruth. (a) Input: Text bounding polygon annotation, defined here by $K=10$ vertices. (b) This polygon consists of $\frac{K-2}{2}=4$ quadrilaterals (shown in different colors).
(c) Each quadrilateral is divided into 
$m$ equal parts, here $m=3$. The Text Center Points (TCP) are marked as red dots, and 
the Text Skeleton (TS) is drawn in orange and coffee colors
 Using knowledge from mathematical geometry, we can get original text annotation points expanded to $\sigma = K + (m - 1)(K - 2)$ points, here 26. 
(d) To focus on the text center region, we delete the two ends of the TS (orange lines). This yields the final TS, here drawn in coffee color. The pink line exemplifies the radius~$R$ in our text representation method (Equation~\ref{dis_equ}). (e) The final heatmap ground truth. The range of the 2D Gaussian kernel is set to $[0.0, 1.0]$.}
\label{pre_fig}
\end{figure}

Our method represents each word or set of words in an image as an arbitrary-length text skeleton  surrounded by a fixed-width region whose pixels have values defined by their distance (``radius'') to the skeleton (Figure~\ref{pre_fig}e).  
 This ``heatmap'' representation for text is sufficiently flexible to represent both straight and curved text.  

The way we generate heatmaps was inspired by previous work \cite{craft,pose_estimation_16}. Instead of simply marking the pixels of the text region as 1 and the background pixels as 0 (e.g., \cite{textsnake,east}), our method assigns a probability to each pixel position in the feature map, indicating the probability that this pixel belongs to the text region (Figure~\ref{pre_fig}e). Naturally, the closer the pixel is to the center of the text, the closer the probability is to 1, and the farther the pixel is to the center of the text, the closer the probability is to 0.

\textbf{Text Skeleton and Radius.}  Each annotated text polygon is defined by $K$ vertices, where $K$ is an even number. First, we use a skeleton to represent each polygon.  We expand the original number of center points on the skeleton to $\sigma = K + (m - 1) \times (K - 2)$ points, where $m$ is a positive integer (see Figure~\ref{pre_fig}b and~\ref{pre_fig}c for more details). In our experiments, $m$ is set to 5. We then pair the coordinates of the upper part of the vertices of the polygon with the coordinates of the lower part of the vertices of the polygon. This yields the following pairs of points: $(P_{up}^{(1)}, P_{down}^{(1)}), (P_{up}^{(2)}, P_{down}^{(2)}),..., (P_{up}^{(\frac{\sigma}{2})}, P_{down}^{(\frac{\sigma}{2})})$. For the $i$th pair of points, we compute the center points as follows:
\begin{equation}
P_{center}^{(i)}=\frac{P_{up}^{(i)} + P_{down}^{(i)}}{2}.
\end{equation}
Then, $P_{center}^{(1)}, P_{center}^{(2)},..., P_{center}^{(\frac{\sigma}{2})}$ are defined as ``Text Center Points (TCP).'' The TCPs are essential for building the ``Text Skeleton (TS).'' We delete the two end groups of TCPs, so that the TCP set is changed from $\left\{ P_{center}^{(1)}, P_{center}^{(2)},..., P_{center}^{(\frac{\sigma}{2})}\right\}$ to $\left\{ P_{center}^{(3)}, P_{center}^{(4)},..., P_{center}^{(\frac{\sigma}{2}-2)}\right\}$. Connecting the center points in the TCP set, then we compute the final polygon skeleton. For each point in the TCP set, we also need their ``Radius (R).'' For the $i$th pair of points, $R^{(i)}$ is defined as follows:
\begin{equation}
\label{dis_equ}
R^{(i)} = \frac{dis(P_{center}^{(i)}, P_{up}^{(i)}) + dis(P_{center}^{(i)}, P_{down}^{(i)})}{2},
\end{equation}
where function $dis(A, B)$ is the Euclidean distance between $A$ and $B$. 

\textbf{2D Gaussian Heatmap Ground Truth Representation.} First we need to compute every point of TS using the Bresenham algorithm \cite{bresenham}, which yields the ``All Text Skeleton Points Set (ATSPS)''. Given the set $R$, which is $\left\{ R^{(3)}, R^{(4)},...R^{(\frac{\sigma}{2}-2)}\right\}$, and a 2D Gaussian kernel, we can compute the heatmap representation for each text bounding polygon. In addition, the range of the values of the generated Heatmap Text Region Groundtruth is set to $[0.0, 1.0]$. These are then used as training labels for the UHT-Net.

Since  the generated heatmap groundtruth is the text skeleton convolved with several 2D Gaussian kernels, each text region is proportional to the length of its text skeleton. So due to our pre-processing algorithm, we suggest that UHT has the potential to be more accurate than other methods when detecting long text regions. 

\subsection{UHT-Net Architecture and Training Objectives}

UHT-Net is a UNet-based \cite{unet} network that predicts score heatmaps of text regions. First, images are contracted to different feature maps. In the expanding process, our method employs either VGG-16 \cite{vgg} or ResNet-50 \cite{resnet} as backbone networks. Then these feature maps are gradually bilinearly expanded to the original size and mixed with the corresponding output of the previous stage in order to accurately detect text of different sizes.

UHT-Net uses an end-to-end training strategy. The definition of the loss function is
\begin{equation}
    L = L_{reg}+\lambda_{1}L_{center} +\lambda_{2}L_{region},
\end{equation}
where $\lambda_{1}$ and $\lambda_{2}$ are both set to 1.0. 

We define $L_{reg}$ as the weighted MSE-Loss (because the ratio between positive and negative samples is unbalanced in the scene text detection datasets), which is defined for an input image $\chi$ to be:
\begin{small}
\begin{equation}
\begin{aligned}
    L_{reg}=\frac{\sum_{text}}{\sum_{text}+\sum_{BG}}\left ( Y_{BG}-f_{\theta}(\chi_{BG}) \right )^2\\
    + \frac{\sum_{BG}}{\sum_{text}+\sum_{BG}}\left ( Y_{text}-f_{\theta}(\chi_{text}) \right )^2,
\end{aligned}
\end{equation}
\end{small}
\noindent
where {\em text} denotes the positive pixels in the heatmap, {\em BG} denotes the negative pixels in the heatmap, $\sum_{text}$ denotes the total number of positive pixels in the heatmap, $\sum_{BG}$ denotes the total number of negative pixels in the heatmap, $Y$ means pixels in the groundtruth heatmap generated by the pre-process, and $f_{\theta}(\chi)$ means pixels in the output of the UHT-Net, where $\theta$ are parameters in the UHT-Net. 

\begin{algorithm}[t]
\label{tfa}
  \begin{algorithmic}[1]
    \State \textbf{Input:} Output heatmap $H$ from UHT-Net; thresholds $T_{top}$, $T_{end}$.
    \State \textbf{\/// Extract center points for each text region:}
    \State Set pixel values in regions where heatmap $H$ pixel values $> T_{top}$  to 1.0, otherwise to 0.0. These regions are defined as {\em CR.}
    \State Find center points {\em CP} for each {\em CR.} 
    \State \textbf{\/// Extract text region:}
    \State $V$ = []
    \For{each {\em CP}}
    \State Initialize zero-valued canvas~$A$ with same shape as heatmap $H$.
    \State stack=set($A[x][y]$).
    \While{{stack}}
    \State $x$, $y$ = stack.pop()
    \If {$judgeFlow$($x-1$, $y$, $x$, $y$)}
    \State stack.add(($x-1$, $y$))
    \EndIf
    \If {$judgeFlow$($x+1$, $y$, $x$, $y$)}
    \State stack.add(($x+1$, $y$))
    \EndIf
    \If {$judgeFlow$($x$, $y-1$, $x$, $y$)}
    \State stack.add(($x$, $y-1$))
    \EndIf
    \If {$judgeFlow$($x$, $y+1$, $x$, $y$)}
    \State stack.add(($x$, $y+1$))
    \EndIf
    \EndWhile
    \State $A$[stack] = 1.0
    \State $C$ = $findCoutour$($A$)
    \State $V$.append($contourExpand$($C$))
    \EndFor
    \State \textbf{Output:} Polygon vertices $V$
  \end{algorithmic}
  \caption{Textfill Algorithm}
  \label{Textfill}
\end{algorithm}

The dice loss \cite{diceloss} for the {\em text center} and {\em text region} is denoted by $L_{center}$ and $L_{region}$   respectively. The {\em text center} is defined as the text region pixels in the output of the UHT-Net and generated groundtruth heatmap pixels that are higher than 0.9. The {\em text region} is defined as the text region pixels in the output of the UHT-Net and generated groundtruth heatmap pixels that are higher than 0.05, which are:
\begin{equation}
    L_{center}=1-\frac{2\left | f_{\theta }(\chi_{center}) \bigcap Y_{center} \right |}{\left | f_{\theta }(\chi_{center}) \right | + \left | Y_{center}  \right |},
\end{equation}
\begin{equation}
    L_{region}=1-\frac{2\left | f_{\theta }(\chi_{region}) \bigcap Y_{region} \right |}{\left | f_{\theta }(\chi_{region}) \right | + \left | Y_{region}  \right |}.
\end{equation}


\subsection{Post-processing: Textfill Algorithm}

Extracting the final predicted text bounding polygons from the output of the UHT-Net is accomplished by our novel post-processing method, the Textfill Algorithm, which is inspired by the floodfill algorithm~\cite{floodfill}. Details are shown in Algorithm~\ref{Textfill}, which uses computer vision tools that can be found in OpenCV. The function $judgeFlow(x_{1}, y_{1}, x_{2}, y_{2})$ is used to expand {\em CR} in Algorithm~\ref{Textfill} to compute the complete text bounding polygon. The definition of $CR$ is at Line 3 of Algorithm~\ref{Textfill}. Function $judgeFlow(x_{1}, y_{1}, x_{2}, y_{2})$ returns {\em true} if $H[x_{1}][y_{1}]<=H[x_{2}][y_{2}]$ and $H[x_{1}][y_{1}]>T_{end}$, or return true if $H[x_{1}][y_{1}]>=T_{end}/2$, where the $H[x][y]$ denotes the pixel in the output from the UHT-Net. Function $contourExpand$ is defined as a dilating process (see morphology tools in OpenCV) with the following kernel:
$$k=
\begin{cases}
8+\frac{S}{750} & \text{$S\in (0, 2 \times 10^4]$}\\
35 & \text{$S>2 \times 10^4$},
\end{cases}$$
where $S$ is the pixel area of the polygonal region $A$.

\section{Experiments}

In this section\footnote{In tables of this section, \textbf{UHT V16} denotes UHT with VGG-16 backbone; \textbf{UHT R50} denotes UHT with ResNet-50 backbone.}, we introduce details of our experiments, including the datasets we use and our training strategy, and provide experimental results and their analysis.

\subsection{Text Detection Datasets Used in Experiments}

\textbf{SynthText} \cite{synthtext} is a large scale dataset with 800k synthetic images that are created by adding English oriented straight text with random fonts, sizes, colors, and orientations to natural images. These synthetic images are quite similar to natural scene images with text.

\textbf{Total-Text} \cite{total-text} is a dataset with images that contain oriented straight and curved text and whose labels are annotated by bounding polygons. The image backgrounds are quite similar to real scenes. This dataset contains 1,255 training and 300 testing images.

\textbf{SCUT-CTW1500} \cite{scut-ctw1500} is another text detection dataset which includes both English and Chinese scripts. SCUT-CTW1500 contains 1,000 training and 500 testing images in which text shape is arbitrary (as for Total-Text). Each text annotation is marked as polygon containing 14 points.

\textbf{MSRA-TD500} \cite{msra-td500} focuses on multilingual oriented straight text in natural scenes. It contains 500 images with English and Chinese scripts, which are split into 300 training and 200 testing images. Text region annotations are marked as rotated rectangles.

\textbf{COCO-Text} \cite{cocotext} is one of the challenges of ICDAR 2017 Robust Reading Competition. Its text instances in the images are English straight text regions distributed in various orientations. It contains 63,686 images in total. 

\subsection{Implementation Details}

\textbf{Data Augmentation.}
The process of training UHT-Net can be divided into two steps: 1) Pretraining with the SynthText dataset, and 2) fine-tuning using the Total-Text, SCUT-CTW1500, MSRA-TD500 or COCO-Text datasets respectively. To further improve  training, we randomly rotated the training images and cropped with areas ranging from 0.24 to 1.0 and aspect ratios ranging from 0.33 to 3. Data augmentation is implemented in both pretraining and fine-tuning processes.

\textbf{Training Strategy of UHT-Net.}
Our methodology was implemented in Pytorch 1.0.1 \cite{pytorch}. UHT-Net was pre-trained on SynthText with one epoch and then fine-tuned using Total-Text, SCUT-CTW1500, MSRA-TD500, or COCO-Text. We adopted the Adam optimizer~\cite{adam} as the learning rate scheme. In the pretraining process, inspired by Smith~\cite{cyc_lr}, we set the initial learning rate to $3\times10^{-5}$ for VGG-16 based UHT and $10^{-4}$ for ResNet-50 based UHT. We did not change it during the pretraining process. In the fine-tuning process, except for COCO-Text, we set the initial learning rate to $10^{-4}$ (the fine-tuning learning rate for COCO-Text is set to $5\times10^{-4}$). The decay rate was 0.8 every 10 epochs. Single-scale training was used. In the pre-training and fine-tuning training processes, we set the batch size to 8 on a single RTX-2080Ti GPU. In the evaluation process, the batch size was set to 1 on a single RTX-2080Ti GPU.

\textbf{Hyperparameters of Textfill Algorithm.}
To show gen\-er\-al results across datasets, two sets of thresholds $T_{top}$, $T_{end}$ were tested: (0.7, 0.2) for Total-Text \& SCUT-CTW1500 and (0.75, 0.2) for MSRA-TD500 \& COCO-Text.

\subsection{Experimental Results}

\subsubsection{Results on Curved Text Detection}
Our results on the benchmarks Total-Text \cite{total-text} and SCUT-CTW1500~\cite{scut-ctw1500} are given in 
Tables~\ref{table_tt} and \ref{table_ctw}, respectively. 
We found that some state-of-the-art methods~\cite{charnet,lomo-ms} included multi-scale testing. To conduct a peer comparison, we ran experiments on curved text detection datasets with single-scale and multi-scale testing (abbreviated as ``MS'' below) separately. Except for the baselines~\cite{total-text, scut-ctw1500}, listed methods without $^{+}$ used the same pre-training and fine-tuning data as we did, otherwise were different.

\textbf{Results on Total-Text Dataset (Table~\ref{table_tt}).} Fine-tuning on Total-Text stops at 300 epochs. During the testing process, each image is set to 700 $\times$ 700. In single-scale testing, our UHT beats all of the state-of-the-art methodologies and keeps the same F-measure score with the newest and most competitive model, CharNet H-88 \cite{charnet}. 
UHT V16 even yields a higher recall rate than all the other state-of-the-art methods, 85.6\%. This indicates that UHT is able to detect text that is missed by other methods.

\begin{figure}[t]
\centering
\centering
\includegraphics[width=1\columnwidth]{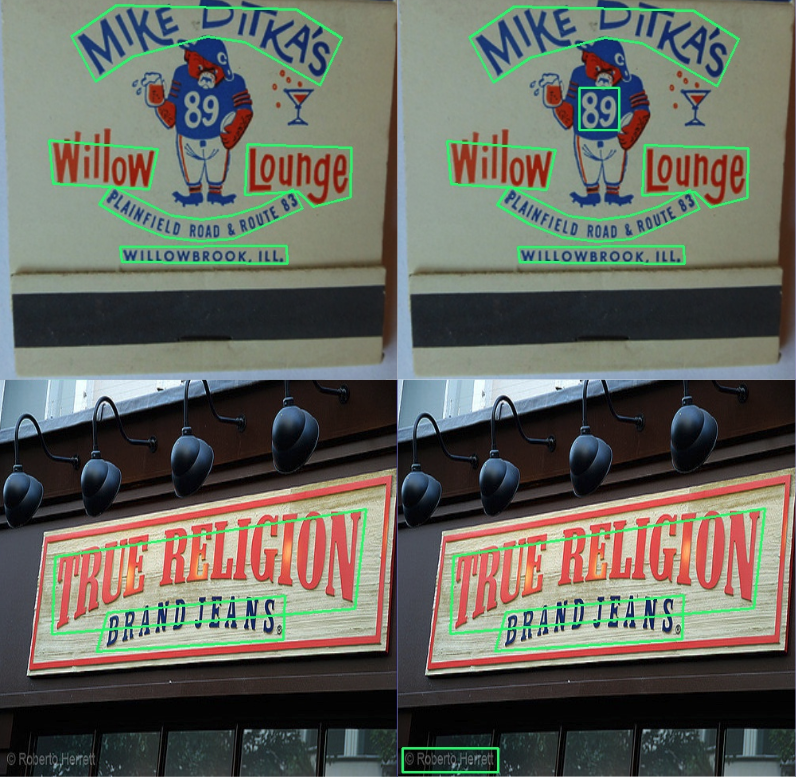}
\caption{Examples of updated ground truth annotations: Left: The groundtruth annotation in
SCUT-CTW1500 missed ``89'' (top) and ``\copyright Roberto Herrett'' (bottom). Right:  Our updated annotations.}
\label{gt_change}
\end{figure}

\textbf{Results on SCUT-CTW1500 Dataset (Table~\ref{table_ctw})} Fine-tuning on SCUT-CTW1500 stops at 307 epochs for UHT V16 and 300 epochs for UHT R50. During the testing process, each image is set to 512 $\times$ 512 because the average size of images in SCUT-CTW1500 is relatively smaller than that of Total-Text. 
Experimental results show that UHT also performs well on SCUT-CTW1500. Surprisingly, UHT found image text that did not appear in the ground truth annotation (see Figure~\ref{gt_change}). We fixed the SCUT-CTW1500 ground truth to include missed words.  To ensure fairness in evaluation, we ran experiments on two different versions of text annotations on the SCUT-CTW1500 dataset, with and without updated ground truth (Table~\ref{table_ctw}). After the ground truth was updated, the recall score of UHT is almost unchanged but the precision score improves a little. We welcome other researchers to run experiments on the updated SCUT-CTW1500 and therefore make it publicly available, see http://www.cs.bu.edu/faculty/betke/UHT.

\begin{table}[]
\centering
\scalebox{0.8}{
\begin{tabular}{ccccc}
\multicolumn{1}{c|}{Methodology} & \multicolumn{1}{c|}{Venue} & \multicolumn{1}{c|}{P (\%)} & \multicolumn{1}{c|}{R (\%)} & F (\%) \\ \hline
\multicolumn{5}{c}{Single-scale Testing} \\ \hline
\multicolumn{1}{c|}{Poly-FRCNN-3 \cite{total-text}} & \multicolumn{1}{c|}{IJDAR-2019} & \multicolumn{1}{c|}{78.0} & \multicolumn{1}{c|}{68.0} & 73.0 \\
\multicolumn{1}{c|}{TextSnake \cite{textsnake}} & \multicolumn{1}{c|}{ECCV-2018} & \multicolumn{1}{c|}{82.7} & \multicolumn{1}{c|}{74.5} & 78.4 \\
\multicolumn{1}{c|}{CSE$^{+}$ \cite{cse}} & \multicolumn{1}{c|}{CVPR-2019} & \multicolumn{1}{c|}{81.4} & \multicolumn{1}{c|}{79.7} & 80.2 \\
\multicolumn{1}{c|}{TextField \cite{textfield}} & \multicolumn{1}{c|}{TIP-2019} & \multicolumn{1}{c|}{81.2} & \multicolumn{1}{c|}{79.9} & 80.6 \\
\multicolumn{1}{c|}{PSENet-1s$^{+}$ \cite{psenet-1s}} & \multicolumn{1}{c|}{CVPR-2019} & \multicolumn{1}{c|}{84.02} & \multicolumn{1}{c|}{77.96} & 80.87 \\
\multicolumn{1}{c|}{FTSN \cite{ftsn}} & \multicolumn{1}{c|}{ICPR-2018} & \multicolumn{1}{c|}{84.7} & \multicolumn{1}{c|}{78.0} & 81.3 \\
\multicolumn{1}{c|}{ICG \cite{icg}} & \multicolumn{1}{c|}{PR-2019} & \multicolumn{1}{c|}{82.9} & \multicolumn{1}{c|}{80.9} & 81.5 \\
\multicolumn{1}{c|}{LOMO \cite{lomo-ms}} & \multicolumn{1}{c|}{CVPR-2019} & \multicolumn{1}{c|}{88.6} & \multicolumn{1}{c|}{75.7} & 81.6 \\
\multicolumn{1}{c|}{CRAFT$^{+}$ \cite{craft}} & \multicolumn{1}{c|}{CVPR-2019} & \multicolumn{1}{c|}{87.6} & \multicolumn{1}{c|}{79.9} & 83.6 \\
\multicolumn{1}{c|}{PSENet\_v2 \cite{psenet_v2}} & \multicolumn{1}{c|}{ICCV-2019} & \multicolumn{1}{c|}{89.3} & \multicolumn{1}{c|}{81.0} & 85.0 \\
\multicolumn{1}{c|}{CharNet H-88 \cite{charnet}} & \multicolumn{1}{c|}{ICCV-2019} & \multicolumn{1}{c|}{\textbf{89.9}} & \multicolumn{1}{c|}{81.7} & \textbf{85.6} \\ \hline
\multicolumn{1}{c|}{\textbf{UHT V16 (Ours)}} & \multicolumn{1}{c|}{-} & \multicolumn{1}{c|}{88.8} & \multicolumn{1}{c|}{\textbf{82.6}} & \textbf{85.6} \\
\multicolumn{1}{c|}{\textbf{UHT R50 (Ours)}} & \multicolumn{1}{c|}{-} & \multicolumn{1}{c|}{88.2} & \multicolumn{1}{c|}{81.8} & 84.9 \\
\hline
\multicolumn{5}{c}{Multi-scale Testing} \\ \hline
\multicolumn{1}{c|}{LOMO MS \cite{lomo-ms}} & \multicolumn{1}{c|}{CVPR-2019} & \multicolumn{1}{c|}{87.6} & \multicolumn{1}{c|}{79.3} & 83.3 \\
\multicolumn{1}{l|}{CharNet H-88 MS \cite{charnet}} & \multicolumn{1}{c|}{ICCV-2019} & \multicolumn{1}{c|}{\textbf{88.0}} & \multicolumn{1}{c|}{85.0} & \textbf{86.5} \\ \hline
\multicolumn{1}{c|}{\textbf{UHT V16 MS (Ours)}} & \multicolumn{1}{c|}{-} & \multicolumn{1}{c|}{85.0} & \multicolumn{1}{c|}{\textbf{85.6}} & 85.3 \\
\multicolumn{1}{c|}{\textbf{UHT R50 MS (Ours)}} & \multicolumn{1}{c|}{-} & \multicolumn{1}{c|}{85.4} & \multicolumn{1}{c|}{84.2} & 84.8
\end{tabular}}
\vspace{0.2cm}
\caption{Experimental results on the Total-Text dataset. 
``P'' means Precision, ``R''  Recall, ``F'' F-measure, $^{*}$ denotes results on updated groundtruth annotations, and  ``MS'' multi-scale testing. }
\label{table_tt}
\end{table}

\begin{table}[]
\centering
\scalebox{0.8}{
\begin{tabular}{ccccc}
\multicolumn{1}{c|}{Methodology} & \multicolumn{1}{c|}{Venue} & \multicolumn{1}{c|}{P (\%)} & \multicolumn{1}{c|}{R (\%)} & F (\%) \\ \hline
\multicolumn{5}{c}{Single-scale Testing} \\ \hline
\multicolumn{1}{c|}{CTD \cite{scut-ctw1500}} & \multicolumn{1}{c|}{PR-2019} & \multicolumn{1}{c|}{74.3} & \multicolumn{1}{c|}{65.2} & 69.5 \\
\multicolumn{1}{c|}{CTD+TLOC \cite{scut-ctw1500}} & \multicolumn{1}{c|}{PR-2019} & \multicolumn{1}{c|}{74.3} & \multicolumn{1}{c|}{69.8} & 73.4 \\
\multicolumn{1}{c|}{TextSnake \cite{textsnake}} & \multicolumn{1}{c|}{ECCV-2018} & \multicolumn{1}{c|}{67.9} & \multicolumn{1}{c|}{\textbf{85.3}} & 75.6 \\
\multicolumn{1}{c|}{CSE$^{+}$ \cite{cse}} & \multicolumn{1}{c|}{CVPR-2019} & \multicolumn{1}{c|}{78.7} & \multicolumn{1}{c|}{76.1} & 77.4 \\
\multicolumn{1}{c|}{LOMO \cite{lomo-ms}} & \multicolumn{1}{c|}{CVPR-2019} & \multicolumn{1}{c|}{\textbf{89.2}} & \multicolumn{1}{c|}{69.6} & 78.4 \\
\multicolumn{1}{c|}{ICG \cite{icg}} & \multicolumn{1}{c|}{PR-2019} & \multicolumn{1}{c|}{82.8} & \multicolumn{1}{c|}{79.8} & 81.3 \\
\multicolumn{1}{c|}{TextField \cite{textfield}} & \multicolumn{1}{c|}{TIP-2019} & \multicolumn{1}{c|}{83.0} & \multicolumn{1}{c|}{79.8} & 81.4 \\
\multicolumn{1}{c|}{CRAFT \cite{craft}} & \multicolumn{1}{c|}{CVPR-2019} & \multicolumn{1}{c|}{86.0} & \multicolumn{1}{c|}{81.1} & 83.5 \\
\multicolumn{1}{c|}{PSENet\_v2 \cite{psenet_v2}} & \multicolumn{1}{c|}{ICCV-2019} & \multicolumn{1}{c|}{86.4} & \multicolumn{1}{c|}{81.2} & 83.7 \\
\multicolumn{1}{c|}{PAN Mask R-CNN$^{+}$ \cite{pan_mask_rcnn}} & \multicolumn{1}{c|}{WACV-2019} & \multicolumn{1}{c|}{86.8} & \multicolumn{1}{c|}{83.2} & 85.0 \\ \hline
\multicolumn{1}{c|}{\textbf{UHT V16 (Ours)}} & \multicolumn{1}{c|}{-} & \multicolumn{1}{c|}{84.3} & \multicolumn{1}{c|}{84.8} & 84.5 \\
\multicolumn{1}{c|}{\textbf{UHT V16* (Ours)}} & \multicolumn{1}{c|}{-} & \multicolumn{1}{c|}{86.2} & \multicolumn{1}{c|}{84.1} & \textbf{85.2} \\
\multicolumn{1}{c|}{\textbf{UHT R50 (Ours)}} & \multicolumn{1}{c|}{-} & \multicolumn{1}{c|}{85.9} & \multicolumn{1}{c|}{83.3} & 84.6 \\
\multicolumn{1}{c|}{\textbf{UHT R50* (Ours)}} & \multicolumn{1}{c|}{-} & \multicolumn{1}{c|}{87.4} & \multicolumn{1}{c|}{82.3} & 84.8 \\ \hline
\multicolumn{5}{c}{Multi-scale Testing} \\ \hline
\multicolumn{1}{c|}{LOMO MS \cite{lomo-ms}} & \multicolumn{1}{c|}{CVPR-2019} & \multicolumn{1}{c|}{\textbf{85.7}} & \multicolumn{1}{c|}{76.5} & 80.8 \\ \hline
\multicolumn{1}{c|}{\textbf{UHT V16 MS (Ours)}} & \multicolumn{1}{c|}{-} & \multicolumn{1}{c|}{83.3} & \multicolumn{1}{c|}{85.4} & 84.4 \\
\multicolumn{1}{c|}{\textbf{UHT V16 MS* (Ours)}} & \multicolumn{1}{c|}{-} & \multicolumn{1}{c|}{85.2} & \multicolumn{1}{c|}{84.7} & \textbf{85.0} \\
\multicolumn{1}{c|}{\textbf{UHT R50 MS (Ours)}} & \multicolumn{1}{c|}{-} & \multicolumn{1}{c|}{81.9} & \multicolumn{1}{c|}{\textbf{86.1}} & 84.0 \\
\multicolumn{1}{c|}{\textbf{UHT R50 MS* (Ours)}} & \multicolumn{1}{c|}{-} & \multicolumn{1}{c|}{84.0} & \multicolumn{1}{c|}{85.5} & 84.7
\end{tabular}}
\vspace{0.2cm}
\caption{Experimental results on the SCUT-CTW1500 dataset: 
``P'' means Precision, ``R''  Recall, ``F'' F-measure, $^{*}$ denotes results on updated groundtruth annotations, and  ``MS'' multi-scale testing.} 
\label{table_ctw}
\end{table}

\textbf{Analysis on Multi-scale Testing}. We tested our model with images of different sizes
(500$\times$500, 700$\times$700, and 900$\times$900), relying on the Fast NMS algorithm to screen out excess text bounding polygons. A reason for the increase of the recall score of multi-scale compared to single-scale testing may be that multi-scale testing combines image information of different sizes, making it easier for UHT to detect text regions that are difficult to detect in single-scale testing. However, false-positive samples are more likely to appear in multi-scale testing. We think that might be caused by the effort of the Fast NMS algorithm or text detection effect on very large images, all of which cause decrease of precision score of multi-scale testing (see Tables~\ref{table_tt} and \ref{table_ctw}).

\subsubsection{Results for Oriented Straight Text Detection}

\textbf{Results on MSRA-TD500 Dataset (Table~\ref{table_td}).} Fine-tuning on MSRA-TD500 stops at 200 epochs. During the testing process, each image is set to 512 $\times$ 512. For UHT, only single-scale texting is implemented in this experiment. The results show that our UHT beats most of the state-of-the-art methods on MSRA-TD500.

\textbf{Results on COCO-Text Dataset (Table~\ref{table_coco}).} 
With the help of the official COCO-Text API, we extracted 15,124 training images and 3,346 validation images, all of which contain at least one text region. However, testing images cannot be extracted. So in the experiments with COCO-Text, experimental results are given based on tests on the validation images for state-of-the-art frameworks and UHT. Fine-tuning on COCO-Text stops at 300 epochs. During the testing process, each image is set to 768 $\times$ 768. For UHT, only single-scale testing is implemented in this experiment. The results (Table~\ref{table_coco}) show that our UHT beats state-of-the-art methods when fine-tuned on COCO-Text training images. For more details of the fine-tuning experiments on IC13 and IC17-MLT datasets, please refer to Section \ref{cha_ga}.


\begin{table}[]
\centering
\scalebox{0.8}{
\begin{tabular}{c|c|c|c|c}
Methodology & Venue & P (\%) & R (\%) & F (\%) \\ \hline
Zhang et al. \cite{DBLP:journals/corr/ZhangZSYLB16} & CVPR-2016 & 83 & 67 & 74 \\
He et al. \cite{DBLP:journals/corr/HeZYL17} & CVPR-2017 & 77 & 70 & 74 \\
EAST$^{\dagger}$ \cite{east} & CVPR-2017 & 87.3 & 67.4 & 76.1 \\
SegLink \cite{seglink} & CVPR-2017 & 86 & 70 & 77 \\
PixelLink$^{\dagger}$ \cite{pixellink} & AAAI-2018 & 83.0 & 73.2 & 77.8 \\
TextSnake \cite{textsnake} & ECCV-2018 & 83.2 & 73.9 & 78.3 \\
RRD$^{\dagger}$ \cite{rrd} & CVPR-2018 & 87 & 73 & 79 \\
Lyu et al.$^{\dagger}$ \cite{DBLP:journals/corr/abs-1802-08948} & CVPR-2018 & 87.6 & 76.2 & 81.5 \\
CRAFT \cite{craft} & CVPR-2019 & \textbf{88.2} & \textbf{78.2} & \textbf{82.9} \\ \hline
\textbf{UHT V16 (Ours)} & - & 84.2 & 76.2 & 80.0 \\
\textbf{UHT R50 (Ours)} & - & 83.2 & 77.0 & 80.0 \\
\end{tabular}}
\vspace{0.1cm}
\caption{Experimental results on MSRA-TD500;
``P'' means Precision, ``R''  Recall, ``F'' F-measure. $^{\dagger}$ denotes multi-scale testing. In this table, training data and testing scale of different methods may not be the same, which inadvertently hinders comparison.}
\label{table_td} 
\end{table}

\begin{table}[]
\centering
\scalebox{0.8}{
\begin{tabular}{ccccc}
\multicolumn{1}{c|}{Methodology} & \multicolumn{1}{c|}{Venue} & \multicolumn{1}{c|}{P (\%)} & \multicolumn{1}{c|}{R (\%)} & F (\%) \\ \hline
\multicolumn{5}{c}{Fine-tuned using COCO-Text} \\ \hline
\multicolumn{1}{c|}{TextSnake\cite{textsnake}} & \multicolumn{1}{c|}{ECCV-2018} & \multicolumn{1}{c|}{54.7} & \multicolumn{1}{c|}{36.3} & 43.6 \\ \hline
\multicolumn{1}{c|}{\textbf{UHT V16 (Ours)}} & \multicolumn{1}{c|}{-} & \multicolumn{1}{c|}{\textbf{62.2}} & \multicolumn{1}{c|}{47.7} & 54.0 \\ 
\multicolumn{1}{c|}{\textbf{UHT R50 (Ours)}} & \multicolumn{1}{c|}{-} & \multicolumn{1}{c|}{60.8} & \multicolumn{1}{c|}{\textbf{49.0}} & \textbf{54.2} \\ \hline
\multicolumn{5}{c}{Fine-tuned using IC13 and IC17-MLT} \\ \hline
\multicolumn{1}{c|}{TextSnake\cite{textsnake}} & \multicolumn{1}{c|}{ECCV-2018} & \multicolumn{1}{c|}{35.3} & \multicolumn{1}{c|}{33.7} & 34.5 \\
\multicolumn{1}{c|}{CRAFT\cite{craft}} & \multicolumn{1}{c|}{CVPR-2019} & \multicolumn{1}{c|}{44.0} & \multicolumn{1}{c|}{28.9} & 34.9 \\ \hline
\multicolumn{1}{c|}{\textbf{UHT V16 (Ours)}} & \multicolumn{1}{c|}{-} & \multicolumn{1}{c|}{43.8} & \multicolumn{1}{c|}{\textbf{43.1}} & 43.4 \\
\multicolumn{1}{c|}{\textbf{UHT R50 (Ours)}} & \multicolumn{1}{c|}{-} & \multicolumn{1}{c|}{\textbf{46.4}} & \multicolumn{1}{c|}{41.5} & \textbf{43.8}
\end{tabular}}
\vspace{0.1cm}
\caption{Experimental results on COCO-Text; ``P'' means Precision, ``R''  Recall, ``F'' F-measure. As far as we know, no new work conducted experiments on testing datasets of COCO-Text since 2019. So in this table, instead of copying directly from other papers, experimental results from state-of-the-art methodologies are reimplemented by us using their official code \cite{craft} and \cite{textsnake_pytorch_website}.} 
\label{table_coco} 
\end{table}

\subsubsection{Generalization Ability}
\label{cha_ga}

A powerful text detection framework should have good generalization ability instead of just overfitting to a particular dataset and reaching high evaluation scores for that dataset. To further verify the generalization ability of UHT, we conducted two additional experiments: (1) we pre-trained and fine-tuned our model on datasets without curved text, here ICDAR-2015~\cite{icdar2015} with 200 epochs, and then evaluated it on the Total-Text and SCUT-CUW1500 datasets. We chose state-of-the-art baselines which were also only fine-tuned on ICDAR-2015~\cite{icdar2015}.  (2) We fine-tuned TextSnake\cite{textsnake} and UHT using the same fine-tuning data as CRAFT\cite{craft}.     

Experimental results (Table~\ref{table_ga}) show that UHT method performs well on two curved and one oriented straight text datasets, {\em even if it is not fine-tuned on them.} It outperforms all listed state-of-the-art baseline methods. We suggest that the powerful generalization ability of UHT is due to its flexibility in text expression as well as the effectiveness of the Textfill Algorithm in extracting text bounding polygons by making full use of the UHT-Net output.

\begin{table}[]
\centering
\scalebox{0.6}{
\begin{tabular}{cc|c|c|c|c|c|c}
\multicolumn{2}{c|}{Dataset} & \multicolumn{3}{c|}{Total-Text} & \multicolumn{3}{c}{SCUT-CTW1500} \\ \hline
\multicolumn{1}{c|}{Methodology} & Venue & P (\%) & R (\%) & F (\%) & P (\%) & R (\%) & F (\%) \\ \hline
\multicolumn{1}{c|}{SegLink \cite{seglink}} & CVPR-2017 & 35.6 & 33.2 & 34.4 & 33.0 & 28.4 & 30.5 \\
\multicolumn{1}{c|}{EAST \cite{east}} & CVPR-2017 & 49.0 & 43.1 & 45.9 & 46.7 & 37.2 & 41.4 \\
\multicolumn{1}{c|}{PixelLink \cite{pixellink}} & AAAI-2018 & 53.5 & 52.7 & 53.1 & 50.6 & 42.8 & 46.4 \\
\multicolumn{1}{c|}{TextSnake \cite{textsnake}} & ECCV-2018 & 61.5 & 67.9 & 64.6 & 65.4 & 63.4 & 64.4 \\
\multicolumn{1}{c|}{CRAFT \cite{craft}} & CVPR-2019 & 63.0 & \textbf{72.9} & 67.6 & 64.5 & 62.0 & 63.3 \\
\multicolumn{1}{c|}{CRAFT* \cite{craft}} & CVPR-2019 & - & - & - & 65.5 & 61.3 & 63.3 \\ \hline
\multicolumn{1}{c|}{\textbf{UHT V16 (Ours)}} & - & 71.3 & 70.2 & \textbf{70.7} & 73.1 & \textbf{75.3} & 74.2 \\
\multicolumn{1}{c|}{\textbf{UHT V16* (Ours)}} & - & - & - & - & \textbf{74.5} & 74.7 & \textbf{74.6} \\
\multicolumn{1}{c|}{\textbf{UHT R50 (Ours)}} & - & \textbf{72.7} & 65.8 & 69.1 & 73.2 & 74.3 & 73.7 \\
\multicolumn{1}{c|}{\textbf{UHT R50* (Ours)}} & - & - & - & - & \textbf{74.5} & 73.5 & 74.0
\end{tabular}}
\vspace{0.2cm}
\caption{Generalization ability on Total-Text and SCUT-CTW1500 datasets
Results of Seglink~\cite{seglink}, EAST~\cite{east}, PixelLink~\cite{pixellink} and TextSnake~\cite{textsnake} were reported by Long et al.~\cite{textsnake}. 
Results of CRAFT~\cite{craft} are from 
the official CRAFT model~\cite{craft_website}, which we fine-tuned on the ICDAR-2015 dataset.
$^{*}$ denotes results with respect to the annotations that we modified for SCUT-CTW1500. We used single-scale testing.}
\label{table_ga}
\end{table}

\subsection{Experimental Results on Text Spotting}

\begin{figure}[t]
\centering
\centering
\includegraphics[width=1\columnwidth]{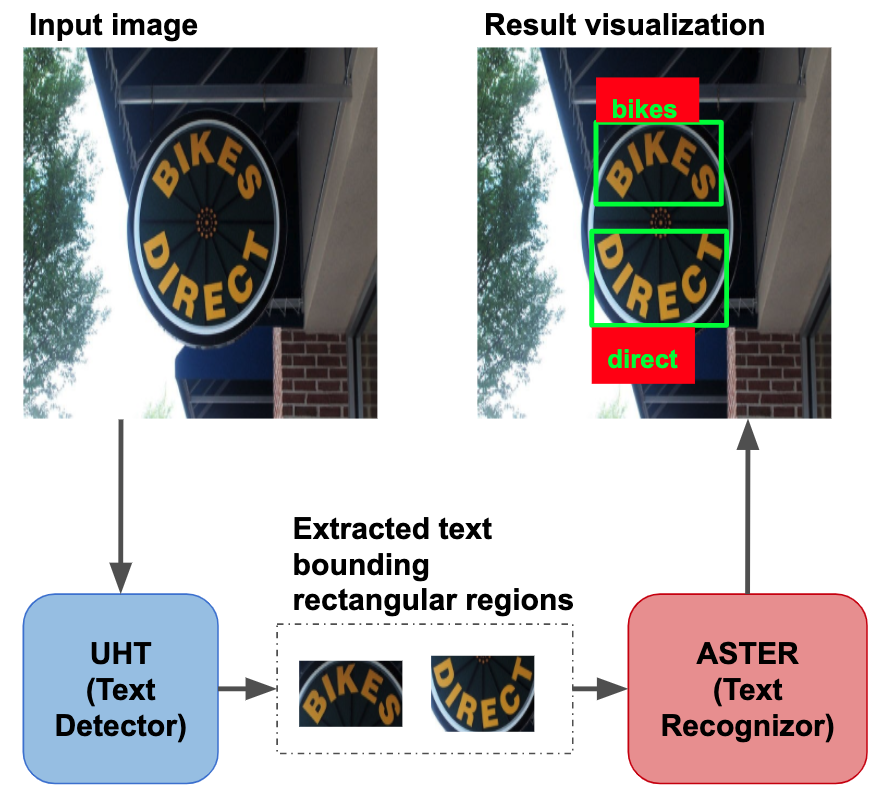}
\caption{Pipeline of the proposed text spotting model UHTA. The model first calls the proposed UHT Detector and then converts UHT's polygonal output regions into horizontally-aligned rectangular regions of text. These text regions are then passed to the state-of-the-art text recognition model ASTER~\cite{aster}, which can accurately recognize the text in these regions and output text strings. So, like a person, UHTA does not only know where the text regions are, but also recognize the content of each text region.}
\label{text_spotting}
\end{figure}


So far, we have shown that the proposed UHT model can accurately localize text in natural scene images. The application range of UHT can be widen when it is embedded into a text recognition framework (Fig.~\ref{text_spotting}). After all, a sighted person would not stop at the task of localizing text but also aim to identify its content.    
To provide a computer vision system who can take on both tasks of detecting and recognizing text in images, we here propose the model UHTA (short for UHT + ASTER~\cite{aster}). 

We now present a peer comparison for UHTA, showing the contribution of UHT in a second text spotting system,  called \textbf{TSA} (TextSnake~\cite{textsnake} + ASTER~\cite{aster}) that applies the pretrained TextSnake model~\cite{textsnake_pytorch_website} with the same training data as UHT. 
The reason why we use ASTER as our text recognizor is that it can efficiently recognize curved and straight text and its code is available. 

We ran experiments using the Total-Text dataset, where annotations of text spotting are included. Single-scale and lexicon-free testing were implemented in the evaluation
for all models.  Experimental results are detailed in Table~\ref{table_spotting}, which show that UHTA has a powerful ability to spot text and outperforms state-of-the-art text spotting systems on the Total-Text dataset. Moreover, since the same conditions were applied for UHTA and TSA, the superior results of UHTA shows the effectiveness of UHT as the text detection module of the text spotting pipeline.


\begin{table}[]
\centering
\scalebox{0.9}{
\begin{tabular}{c|c|c}
Methodology & Venue & F-measure (\%) \\ \hline
Textboxes \cite{textboxes} & AAAI-2017 & 36.3 \\
Mask TextSpotter \cite{mask_textspotter} & ECCV-2018 & 52.9 \\
TextNet \cite{textnet} & ACCV-2018 & 54.0 \\
CharNet H-88 \cite{charnet} & ICCV-2019 & 66.6 \\ \hline
TSA \cite{textsnake, aster} & ECCV-2018 & 58.1 \\
\textbf{UHTA V16 (Ours)} \cite{aster} & - & 75.7 \\
\textbf{UHTA R50 (Ours)} \cite{aster} & - & \textbf{77.6}
\end{tabular}}
\vspace{0.2cm}
\caption{Experimental results of TSA and UHTA on the Total-Text Dataset. Pretrained ASTER~\cite{aster} model is downloaded from official pytorch reimplementation~\cite{aster_website}. Evaluation method for UHTA and TSA is end-to-end recognition from \cite{e2e_website}. Annotations are horizontal text-bounding rectangles because UHT and TextSnake outputs horizontal text-bounding rectangles in UHTA and TSA. 
No distinction between uppercase and lowercase was made when we evaluated UHTA and TSA. The listed F-measures of the prior works were reported in their original papers. UHTA V16 denotes UHTA with VGG-16backbone; UHTA R50 denotes UHTA with ResNet-50 backbone.} 
\label{table_spotting}
\end{table}

\subsection{Analysis and Discussion}

\textbf{Framework Features.} 
UHT is robust to different scales and shapes of text from natural scene images. UHT treats the text in the natural scene images directly as positive regions instead of the composition of different geometry attributes, whether it is oriented straight text or curved text. Textfill algorithm can also flexibly and accurately extract the text in the images according to the output of UHT-Net. Even if text regions are very close to each other, UHT can accurately separate words, outperforming most of the state-of-the-art methodologies in the text detection field.

\textbf{Multilingual Ability.}  SCUT-CTW1500 and MSRA-T\-D500 contain English and Chinese scripts. Our
results show the effectiveness of UHT in detecting scripts in a Latin language like English and Sino-Tibetan language like Chinese.

\textbf{Generalization Ability.} UHT outperforms state-of-the-art text detection frameworks by at least 1.5 percent points in the F-measure of Total-Text~\cite{total-text} (Table~\ref{table_ga}), at least 10.4 percent points in the F-measure of SCUT-CTW1500~\cite{scut-ctw1500} (Table~\ref{table_ga}), and 8.5 percent points in the F-measure of COCO-Text~\cite{cocotext} (Table~\ref{table_coco}), even when not fine-tuned on them.  This shows that UHT not only performs well when fine-tuned to a specific dataset, but also performs well when not.  We thus conclude that UHT is robust and has a strong generalization ability.

\textbf{Speed Analysis.} Table~\ref{tab_speed} reveals that the speed of UHT when dealing with curved text is slower than with oriented straight text. We think this might be caused by the original ground truth representation of the text region. Oriented straight text is represented by a rectangle but curved text by a more complicated multi-vertex polygon. 

\begin{table}[t]
\centering
\scalebox{0.65}{
\begin{tabular}{c|c|c|c|c}
Backbone & Total-Text & SCUT-CTW1500 & MSRA-TD500 & COCO-Text \\ \hline
UHT V16 & 1.6 & 2.1 & 2.6 & 5.2 \\
UHT R50 & 1.9 & 1.8 & 3.7 & 4.5
\end{tabular}}
\vspace{0.1cm}
\caption{Speed of UHT. The unit of measure is FPS.}
\label{tab_speed}
\end{table}


\textbf{The backbone UHT-Net.} Interestingly, analyzing all experimental results, we found that usually UHT R50 performs slightly better than UHT V16; but sometimes the opposite occurs. 
We think this might be caused by the hyperparameter setting: our choice of hyperparameters may not allow UHT V16 or UHT R50 to exert their optimal abilities compared with another model for a particular dataset.

\textbf{Text Spotting Analysis.} The strong experimental results of UHTA (Table~\ref{table_spotting}) show the strength of UHT as an application -- UHT performs excellent when used as a text detector in a pipeline-based text spotting system.

\section{Conclusion}

In this paper, we proposed a new text detection model called UHT that, with little information, can effectively detect text in natural scene images. UHT performs well in experiments with publicly available datasets. This includes experiments when UHT is fined-tuned and tested on a specific dataset and when fine-tuned and tested on different datasets. We fixed ground truth annotation errors of the SCUT-CTW1500 dataset and make the corrected ground truth publicly available. We further showed the scope of UHT by implementing a pipeline-based text spotting system that
improves the results of other state-of-the-art text spotting frameworks by a range of 9.1--41.3 percent points in the F-measure.
In the future, we plan to explore the possibility of detecting scripts of languages other than 
English and Chinese, such as Korean and Arabic, with UHT.


\vspace*{.4cm}

\noindent
{\bf Acknowledgements}

This work has been partially supported by the National Science Foundation, grant 1838193, and the Boston University Hariri Institute for Computing.


{\small
\bibliographystyle{ieee}
\bibliography{egbib}

\begin{thebibliography}{10}\itemsep=-1pt

\bibitem{bresenham}
\url{https://en.wikipedia.org/wiki/Bresenham$\%$27s_line_algorithm}.

\bibitem{floodfill}
\url{https://en.wikipedia.org/wiki/Flood_fill}.

\bibitem{textsnake_pytorch_website}
\url{https://github.com/princewang1994/TextSnake.pytorch}.

\bibitem{craft_website}
\url{https://github.com/clovaai/CRAFT-pytorch}.

\bibitem{aster_website}
\url{https://github.com/ayumiymk/aster.pytorch}.

\bibitem{e2e_website}
\url{https://github.com/liuheng92/OCR_EVALUATION}.

\bibitem{craft}
Y.~Baek, B.~Lee, D.~Han, S.~Yun, and H.~Lee.
\newblock Character region awareness for text detection.
\newblock {\em CoRR}, abs/1904.01941, 2019.

\bibitem{total-text}
C.~K. Chng and C.~S. Chan.
\newblock Total-text: {A} comprehensive dataset for scene text detection and
  recognition.
\newblock {\em CoRR}, abs/1710.10400, 2017.

\bibitem{ftsn}
Y.~Dai, Z.~Huang, Y.~Gao, and K.~Chen.
\newblock Fused text segmentation networks for multi-oriented scene text
  detection.
\newblock {\em CoRR}, abs/1709.03272, 2017.

\bibitem{pixellink}
D.~Deng, H.~Liu, X.~Li, and D.~Cai.
\newblock Pixellink: Detecting scene text via instance segmentation.
\newblock {\em CoRR}, abs/1801.01315, 2018.

\bibitem{swt}
B.~Epshtein, E.~Ofek, and Y.~Wexler.
\newblock Detecting text in natural scenes with stroke width transform.
\newblock {\em 2010 IEEE Computer Society Conference on Computer Vision and
  Pattern Recognition}, pages 2963--2970, 2010.

\bibitem{synthtext}
A.~Gupta, A.~Vedaldi, and A.~Zisserman.
\newblock Synthetic data for text localisation in natural images.
\newblock {\em CoRR}, abs/1604.06646, 2016.

\bibitem{resnet}
K.~He, X.~Zhang, S.~Ren, and J.~Sun.
\newblock Deep residual learning for image recognition.
\newblock {\em CoRR}, abs/1512.03385, 2015.

\bibitem{DBLP:journals/corr/HeZYL17}
W.~He, X.~Zhang, F.~Yin, and C.~Liu.
\newblock Deep direct regression for multi-oriented scene text detection.
\newblock {\em CoRR}, abs/1703.08289, 2017.

\bibitem{pan_mask_rcnn}
Z.~Huang, Z.~Zhong, L.~Sun, and Q.~Huo.
\newblock Mask {R-CNN} with pyramid attention network for scene text detection.
\newblock {\em CoRR}, abs/1811.09058, 2018.

\bibitem{icdar2015}
D.~Karatzas, L.~Gomez-Bigorda, A.~Nicolaou, S.~Ghosh, A.~Bagdanov, M.~Iwamura,
  J.~Matas, L.~Neumann, V.~R. Chandrasekhar, S.~Lu, et~al.
\newblock Icdar 2015 competition on robust reading.
\newblock In {\em 2015 13th International Conference on Document Analysis and
  Recognition (ICDAR)}, pages 1156--1160. IEEE, 2015.

\bibitem{adam}
D.~P. Kingma and J.~Ba.
\newblock Adam: A method for stochastic optimization.
\newblock {\em arXiv preprint arXiv:1412.6980}, 2014.

\bibitem{textboxes}
M.~Liao, B.~Shi, X.~Bai, X.~Wang, and W.~Liu.
\newblock Textboxes: {A} fast text detector with a single deep neural network.
\newblock {\em CoRR}, abs/1611.06779, 2016.

\bibitem{rrd}
M.~Liao, Z.~Zhu, B.~Shi, G.~Xia, and X.~Bai.
\newblock Rotation-sensitive regression for oriented scene text detection.
\newblock {\em CoRR}, abs/1803.05265, 2018.

\bibitem{ssd}
W.~Liu, D.~Anguelov, D.~Erhan, C.~Szegedy, S.~E. Reed, C.~Fu, and A.~C. Berg.
\newblock {SSD:} single shot multibox detector.
\newblock {\em CoRR}, abs/1512.02325, 2015.

\bibitem{scut-ctw1500}
Y.~Liu, L.~Jin, S.~Zhang, and S.~Zhang.
\newblock Detecting curve text in the wild: New dataset and new solution.
\newblock {\em CoRR}, abs/1712.02170, 2017.

\bibitem{cse}
Z.~Liu, G.~Lin, S.~Yang, F.~Liu, W.~Lin, and W.~L. Goh.
\newblock Towards robust curve text detection with conditional spatial
  expansion.
\newblock {\em CoRR}, abs/1903.08836, 2019.

\bibitem{fcn}
J.~Long, E.~Shelhamer, and T.~Darrell.
\newblock Fully convolutional networks for semantic segmentation.
\newblock {\em CoRR}, abs/1411.4038, 2014.

\bibitem{textsnake}
S.~Long, J.~Ruan, W.~Zhang, X.~He, W.~Wu, and C.~Yao.
\newblock Textsnake: A flexible representation for detecting text of arbitrary
  shapes.
\newblock In {\em The European Conference on Computer Vision (ECCV)}, September
  2018.

\bibitem{mask_textspotter}
P.~Lyu, M.~Liao, C.~Yao, W.~Wu, and X.~Bai.
\newblock Mask textspotter: An end-to-end trainable neural network for spotting
  text with arbitrary shapes.
\newblock {\em CoRR}, abs/1807.02242, 2018.

\bibitem{DBLP:journals/corr/abs-1802-08948}
P.~Lyu, C.~Yao, W.~Wu, S.~Yan, and X.~Bai.
\newblock Multi-oriented scene text detection via corner localization and
  region segmentation.
\newblock {\em CoRR}, abs/1802.08948, 2018.

\bibitem{mser}
J.~Matas, O.~Chum, M.~Urban, and T.~Pajdla.
\newblock Robust wide-baseline stereo from maximally stable extremal regions.
\newblock {\em Image and vision computing}, 22(10):761--767, 2004.

\bibitem{pose_estimation_16}
A.~Newell, K.~Yang, and J.~Deng.
\newblock Stacked hourglass networks for human pose estimation.
\newblock {\em CoRR}, abs/1603.06937, 2016.

\bibitem{pytorch}
A.~Paszke, S.~Gross, S.~Chintala, G.~Chanan, E.~Yang, Z.~DeVito, Z.~Lin,
  A.~Desmaison, L.~Antiga, and A.~Lerer.
\newblock Automatic differentiation in pytorch.
\newblock 2017.

\bibitem{faster-rcnn}
S.~Ren, K.~He, R.~B. Girshick, and J.~Sun.
\newblock Faster {R-CNN:} towards real-time object detection with region
  proposal networks.
\newblock {\em CoRR}, abs/1506.01497, 2015.

\bibitem{unet}
O.~Ronneberger, P.~Fischer, and T.~Brox.
\newblock U-net: Convolutional networks for biomedical image segmentation.
\newblock {\em CoRR}, abs/1505.04597, 2015.

\bibitem{seglink}
B.~Shi, X.~Bai, and S.~J. Belongie.
\newblock Detecting oriented text in natural images by linking segments.
\newblock {\em CoRR}, abs/1703.06520, 2017.

\bibitem{aster}
B.~Shi, M.~Yang, X.~Wang, P.~Lyu, C.~Yao, and X.~Bai.
\newblock Aster: An attentional scene text recognizer with flexible
  rectification.
\newblock {\em IEEE transactions on pattern analysis and machine intelligence},
  2018.

\bibitem{vgg}
K.~Simonyan and A.~Zisserman.
\newblock Very deep convolutional networks for large-scale image recognition.
\newblock {\em arXiv preprint arXiv:1409.1556}, 2014.

\bibitem{cyc_lr}
L.~N. Smith.
\newblock No more pesky learning rate guessing games.
\newblock {\em CoRR}, abs/1506.01186, 2015.

\bibitem{diceloss}
C.~H. Sudre, W.~Li, T.~Vercauteren, S.~Ourselin, and M.~J. Cardoso.
\newblock Generalised dice overlap as a deep learning loss function for highly
  unbalanced segmentations.
\newblock {\em CoRR}, abs/1707.03237, 2017.

\bibitem{textnet}
Y.~Sun, C.~Zhang, Z.~Huang, J.~Liu, J.~Han, and E.~Ding.
\newblock Textnet: Irregular text reading from images with an end-to-end
  trainable network.
\newblock {\em CoRR}, abs/1812.09900, 2018.

\bibitem{icg}
J.~Tang, Z.~Yang, Y.~Wang, Q.~Zheng, Y.~Xu, and X.~Bai.
\newblock Detecting dense and arbitrary-shaped scene text by instance-aware
  component grouping.
\newblock {\em Pattern Recognition}, 2019.

\bibitem{cocotext}
A.~Veit, T.~Matera, L.~Neumann, J.~Matas, and S.~Belongie.
\newblock Coco-text: Dataset and benchmark for text detection and recognition
  in natural images.
\newblock In {\em arXiv preprint arXiv:1601.07140}, 2016.

\bibitem{psenet-1s}
W.~Wang, E.~Xie, X.~Li, W.~Hou, T.~Lu, G.~Yu, and S.~Shao.
\newblock Shape robust text detection with progressive scale expansion network.
\newblock {\em CoRR}, abs/1903.12473, 2019.

\bibitem{psenet_v2}
W.~Wang, E.~Xie, X.~Song, Y.~Zang, W.~Wang, T.~Lu, G.~Yu, and C.~Shen.
\newblock Efficient and accurate arbitrary-shaped text detection with pixel
  aggregation network.
\newblock {\em arXiv preprint arXiv:1908.05900}, 2019.

\bibitem{charnet}
L.~Xing, Z.~Tian, W.~Huang, and M.~R. Scott.
\newblock Convolutional character networks.
\newblock {\em arXiv preprint arXiv:1910.07954}, 2019.

\bibitem{textfield}
Y.~Xu, Y.~Wang, W.~Zhou, Y.~Wang, Z.~Yang, and X.~Bai.
\newblock Textfield: Learning {A} deep direction field for irregular scene text
  detection.
\newblock {\em CoRR}, abs/1812.01393, 2018.

\bibitem{msra-td500}
C.~{Yao}, X.~{Bai}, W.~{Liu}, Y.~{Ma}, and Z.~{Tu}.
\newblock Detecting texts of arbitrary orientations in natural images.
\newblock In {\em 2012 IEEE Conference on Computer Vision and Pattern
  Recognition}, pages 1083--1090, June 2012.

\bibitem{DBLP:journals/corr/YaoBSZZC16}
C.~Yao, X.~Bai, N.~Sang, X.~Zhou, S.~Zhou, and Z.~Cao.
\newblock Scene text detection via holistic, multi-channel prediction.
\newblock {\em CoRR}, abs/1606.09002, 2016.

\bibitem{lomo-ms}
C.~Zhang, B.~Liang, Z.~Huang, M.~En, J.~Han, E.~Ding, and X.~Ding.
\newblock Look more than once: An accurate detector for text of arbitrary
  shapes.
\newblock {\em CoRR}, abs/1904.06535, 2019.

\bibitem{DBLP:journals/corr/ZhangZSYLB16}
Z.~Zhang, C.~Zhang, W.~Shen, C.~Yao, W.~Liu, and X.~Bai.
\newblock Multi-oriented text detection with fully convolutional networks.
\newblock {\em CoRR}, abs/1604.04018, 2016.

\bibitem{east}
X.~Zhou, C.~Yao, H.~Wen, Y.~Wang, S.~Zhou, W.~He, and J.~Liang.
\newblock East: An efficient and accurate scene text detector.
\newblock In {\em The IEEE Conference on Computer Vision and Pattern
  Recognition (CVPR)}, July 2017.

\bibitem{textmountaion}
Y.~Zhu and J.~Du.
\newblock Textmountain: Accurate scene text detection via instance
  segmentation.
\newblock {\em CoRR}, abs/1811.12786, 2018.

\end{thebibliography}
}

\end{document}